\documentclass[manuscript]{acmart}

\AtBeginDocument{%
  \providecommand\BibTeX{{%
    \normalfont B\kern-0.5em{\scshape i\kern-0.25em b}\kern-0.8em\TeX}}}

\usepackage{color}
\usepackage{subcaption}
\usepackage{varwidth}
\usepackage{multirow}
\usepackage{bbm}

\begin{document}

\title{Trajectory Based Podcast Recommendation}

\author{Greg Benton}
\email{gwb260@nyu.edu}
\affiliation{%
  \institution{New York University* \thanks{*Work done while at Spotify}}
  \city{New York}
}

\author{Ghazal Fazelnia}
\affiliation{%
  \institution{Spotify}
  \city{New York}
}
\author{Alice Wang}
\affiliation{%
  \institution{Spotify}
  \city{New York}
}
\author{Ben Carterette}
\affiliation{%
  \institution{Spotify}
  \city{New York}
}


\begin{abstract}
 Podcast recommendation is a growing area of research that presents new challenges and opportunities. Individuals interact with podcasts in a way that is distinct from most other media; and primary to our concerns is distinct from music consumption. We show that successful and consistent recommendations can be made by viewing users as moving through the podcast library \emph{sequentially}. Recommendations for future podcasts are then made using the trajectory taken from their sequential behavior. Our experiments provide evidence that user behavior is confined to local trends, and that listening patterns tend to be found over short sequences of similar types of shows. Ultimately, our approach gives a $450\%$ increase in effectiveness over a collaborative filtering baseline.
\end{abstract}



\begin{teaserfigure}
    \centering
  \includegraphics[width=0.7\textwidth]{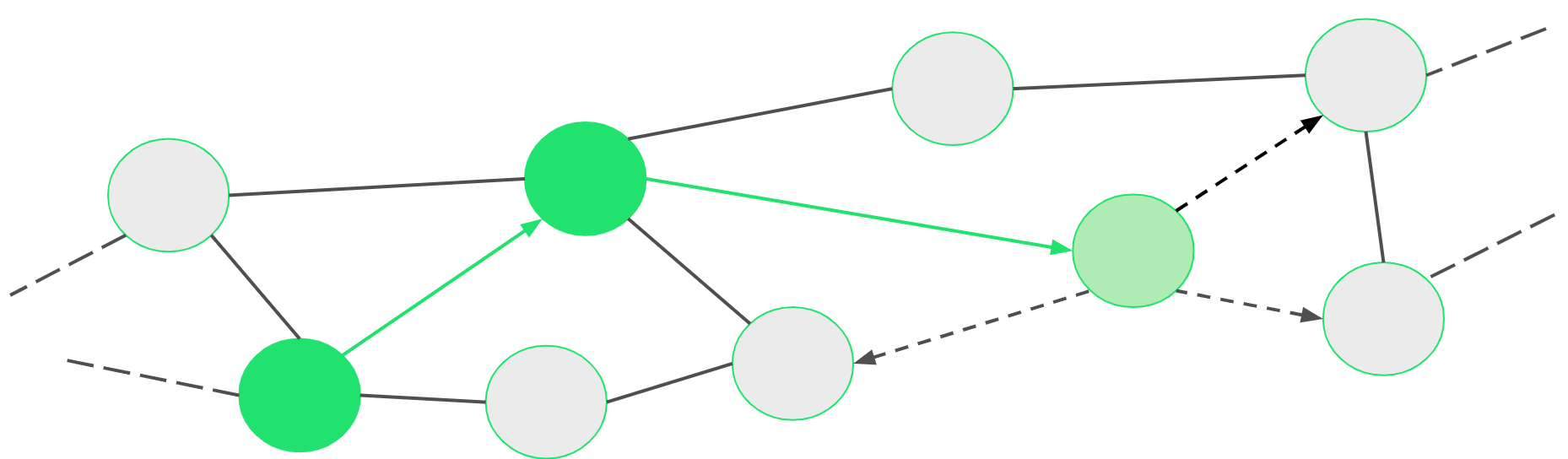}
  \caption{Representation of a user listening to a sequence of podcasts. We view the library of podcast shows as a graph with shows as nodes and edges representing a shared topic between two shows. By coupling a user's history (dark green nodes) with their current position (light green node) we can provide them with recommendations that incorporates a sense of the user's \emph{trajectory} in the graph of podcasts.}
  \label{fig:teaser}
\end{teaserfigure}

\maketitle

\section{Introduction}
Podcasts have seen rapid growth in recent years and have become an essential part of listening habits. In the US alone, 75\% of the population is familiar with the term “podcasting”, 55\% has listened to a podcast, and 37\% has listened to a podcast in the last month. Streaming platforms now serve more than 850,000 active podcasts with over 30 million podcast episodes \citep{EdisonResearchPod2020}.

With such an expansive space of entities, recommendation and search play a crucial role to match users with their desired podcasts. Recommender systems models and algorithms are an integral component of the majority of online markets, streaming platforms, and entertainment and information services. 
We typically observe how a set of users interact with a set of items and train models to learn interests and interaction behaviors. 
Collaborative filtering is among the most widely used approaches that makes recommendations based on learning such patterns among users and items 
\citep{schafer2007collaborative}. Deep learning-based models have emerged as an essential tool for addressing challenges in modeling and predicting complex structures in personalization tasks. They have been successfully applied to learn representations of users and items, model various user behavior and feedback, and make efficient recommendations \citep{zhang2019deep}. In many application domains, user-item interactions occur over time, and it has been shown that incorporating the sequentially-ordered logs can benefit the performance of recommendations \citep{quadrana2018sequence}. 

Prior works on podcast recommendation have focused on content modeling and topic retrieval based textual and non-textual information \citep{tsagkias2010predicting,yang2018understanding}. Recently, \citet{yang2019more} have shown that non-textual aspects of podcasts such as features based on music and speech could improve the performance of topic-based podcast popularity prediction.  In recent years, incorporating a Knowledge Graph into the recommender systems has enhanced the performance by allowing for side information for the entities \citep{khrouf2013hybrid, catherine2016personalized, palumbo2017entity2rec}. Knowledge graphs are heterogeneous graphs with entity nodes and relational edges between them. They have shown promising results for recommendation precision and have resulted in better explainability of recommendations \citep{zhang2016collaborative}.

In this work, we introduce a novel method for podcast recommendation based on sequentially ordered user interactions as well as information from the podcast knowledge graph. By learning user trajectories in discovering podcasts, and including the semantic relationship and embeddings from the knowledge graph we are able to produce precise recommendations. 
To the best of our knowledge, this is the first work that utilizes both sequential interaction as well as content embeddings based on a knowledge graph representation for podcast recommendation. 
Our experimental results show significant improvement in success of recommendations on a real-world dataset. 
In particular, we show that our approach results is $4$ times more precision in podcast recommendation compared to collaborative filtering methods.

\section{Model Design and Baseline Comparisons}\label{sec:modeldesign}

In this section we provide an overview of our method as well as initial comparisons to existing baselines.

\subsection{Recommendation Pipeline}
\label{sec: basics}

Unless otherwise stated, in the sections that follow we apply the same general framework to making recommendations. 
Prior to any modeling we restrict ourselves to the top shows on the platform, such that $90\%$ of all listening in the United States is accounted for.
Recommendations are made at the show level, and we consider a user as having listened to a show based on the first time they streamed an episode of that show for at least $30$ seconds (this time restriction is to discard any accidental plays or intentional skips).

For a given user we take a sequence of shows they have listened to, $s_{t-k}, s_{t-k+1}, \dots, s_{t}$, indexed by consumption order, and make a recommendation for the next show to which they will listen, $s_{t+1}$.
To make these recommendations we first embed the previously heard shows into a vector space generated by taking a DistMult embedding of the knowledge graph of the podcast library.
The knowledge graph matches publicly available data sources (such as Wikipedia) to the descriptions of podcasts \citep{yang2014embedding}.
These embeddings then incorporate high level semantic relationships between shows based on the Wikipedia pages that can be connected with the show descriptions, creating relationships between shows that connect to related entities \cite{yang2014embedding}.
We denote the embedding of show $s_t$ as $s_{e, t}$.

With the embeddings of the shows determined we use a recurrent neural network (RNN) to learn a function $f$ that takes a sequence of embedded shows as input and generates a distribution over candidate shows for the next show to which the user will listen. Specifically we seek $f$ such that,
\begin{equation}
    f(s_{e, t-k}, \dots, s_{e, t}) \rightarrow p(s_{t+1}).
\end{equation}
We leave the embedding and the choice of $k$ out for now, as we give careful consideration to these in Sections \ref{sec: structure} and \ref{sec: sequence-length}.

In the following experiments we define $f$ to be a RNN with three LSTM blocks with $512$ hidden units, followed by two dense feed-forward layers of width $512$ and $1024$, with an output layer that is the size of the shows being considered.
We then apply a softmax to the outputs of the RNN, giving a probability distribution over shows, representing the probability that each show is the next one consumed by a user. We can easily rank our recommendations by simply ranking shows by our estimated probability of them being consumed next by a user. 

\subsection{Initial Comparisons}

Prior to detailing the development and performance of the model introduced in this work, we give baseline comparison against a standard collaborative filtering approach. This section serves primarily as a justification that combining semantically informed embeddings with deep sequential models can provide significant improvement over more standard approaches. 
For this experiment we hold out the last show in each user's listening history as the target, and compare the rankings for these held out shows in the recommendations made. 

We make the collaborative filtering recommendations using a matrix factorization approach, in which we factor a user-show matrix to get embeddings of both the users and shows in a $40$ dimensional feature space.
The user-show matrix, denoted $X$, is a $0-1$ matrix, in which each row corresponds to a user and each column to a show. Entries of the matrix are $1$ if a user listened to a particular show and $0$ otherwise. 

We use the coordinate descent algorithm from \citet{scikit-learn} to find $W$ and $H$ such that $X \approx WH$ \citep{cichocki2009fast}. The rows of $W$ are then embeddings of the users and the columns of $H$ are embeddings of the shows. We then recommend shows for users in decreasing order of the cosine similarity between the user embeddings and the show embeddings.

\begin{figure}
  \begin{subfigure}[b]{0.4\textwidth}
    \includegraphics[width=\textwidth]{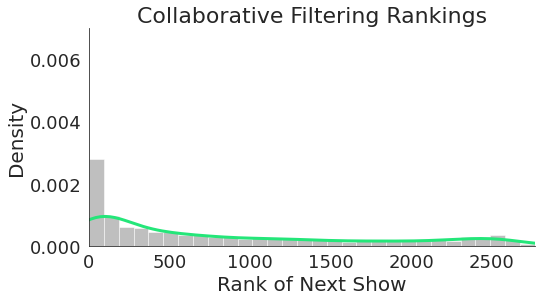}
  \end{subfigure}
  \begin{subfigure}[b]{0.4\textwidth}
    \includegraphics[width=\textwidth]{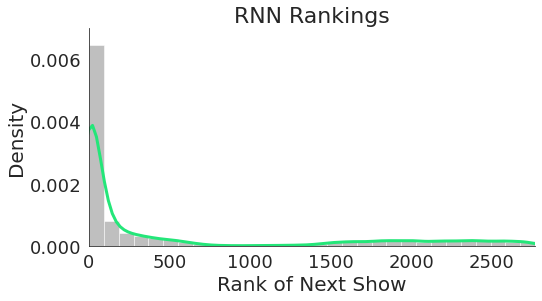}
  \end{subfigure}
  \caption{\textbf{Left}: Ranking of recommendations made by matrix-factorization based collaborative filtering, \textbf{Right}: Ranking of recommendations made by using an RNN in conjunction with the knowledge graph embeddings. The RNN based approach gives an immediate improvement over the collaborative filtering approach, assigning higher ranks to the true next show.}
  \label{fig: collab-filtering}
\end{figure}

Table \ref{table: collab-filtering} and Figure \ref{fig: collab-filtering} show comparisons between the collaborative filtering approach described above and the RNN approach given in Section \ref{sec: basics}. 
In the RNN model we use only the previous two shows a user listened to and attempt to predict the last show in their history. For both methods we take out a training set of $40,000$ users and a test set of $5,000$. 
Table \ref{table: collab-filtering} compares both the success at $20$ and the mean reciprocal rank of the recommendations. Success at $20$ and mean reciprocal rank are defined as, respectively,
\begin{equation}\label{eqn: mrr}
    Sa20 = \frac{1}{N}\sum_{i=1}^{N}\mathbbm{1}[rank_i \leq 20], \qquad
    MRR = \frac{1}{N}\sum_{i=1}^{N}\frac{1}{rank_i},
\end{equation}
where $rank_i$ is the rank of the next show for the $i^{th}$ user in the test set. If we consistently rank the next show high in our recommendations (i.e. $rank_i$ is close to $1$) we get high success and $MRR$, thus we give preference to models with the highest values for both measures.  

\begin{table}[h]
    \centering
    \begin{tabular}{ c  c  c}
         & Collab. Filtering & RNN \\
         \hline \hline
         MRR & $0.0340$ & $\mathbf{0.2069}$\\
         \hline
         Success at $20$ & $0.1186$ & $\mathbf{0.4040}$
    \end{tabular}
    \caption{A comparison on mean reciprocal rank and success at $20$ between the collaborative filtering rankings and the RNN generated rankings.}
    \label{table: collab-filtering}
\end{table}

The density plots in Figure \ref{fig: collab-filtering} show how the true next shows consumed by users were ranked in the recommendations made by each model (the green curve shows a kernel density estimate of the true next show being assigned a given rank). The RNN approach is much more successful at assigning a high rank to the true next show (close to $1$), leading to higher measures of success at $20$ and MRR.

The improvements seen in transitioning from a classical collaborative filtering approach to our knowledge graph based deep model show the promise of our method. Moving forward we continue to outdo these baselines, however for parsimony and clarity we do not compare all the extensions made to our model back to collaborative filtering.

\section{Modeling Choices and Development}\label{sec:datacuration}

Much of the time spent in developing our method is devoted to finding the optimal model setup, and a significant portion of this endeavor is to find the right sequences of podcast consumption for which we can make good recommendations.
We detail the choices of model construction and data selection as follows:
\begin{itemize}
    \item Section \ref{sec: structure}: The inclusion of semantic information (via knowledge graph embeddings) as compared to embeddings based on user behavior, as well as the restriction of recommendations to within certain topics of interest,
    \item Section \ref{sec: sequence-length}: Constraining the sequences of user's history to certain record lengths over shorter time windows
    \item Section \ref{sec: sequential}: Returning to the original modeling choice of employing a RNN and confirming that the sequential structure is needed to make successful recommendations.
\end{itemize}

\subsection{The Importance of Structure}\label{sec: structure}

The space of podcasts is large and diverse, containing many shows of differing styles, topics, and popularity. Experimentally we find that incorporating as much information about the structure of the space of podcasts as possible is core to making accurate and successful recommendations. We can imbue our model with some knowledge of the structure of the data in two ways. The first is by subsetting user listening data based on the topics of the shows to which they have listened. The second is to use embeddings from the knowledge graph rather than embeddings learned from user data. We provide more details on each of these approaches in the next two sections, followed by a comparison across all combinations of methods considered. 

\paragraph{Topic Constraints}

We consider podcasts at the \textit{show} level (as opposed to considering individual episodes) in the context of a show-topic graph. The show topic graph contains nodes representative of each show in the library, with edges between shows if they share a topic. Topics here are coarse labels such as \textit{true crime} or \textit{daily news}, and each show may be associated with any number of topics. When considering sequences of shows a user has consumed we can filter in one of two distinct ways to generate sequences with which to train models for recommendation, in order of least to most constrained these are as follows. 
\begin{itemize}
    \item \textbf{Unconstrained}: we impose no restrictions and maintain the data as just the raw sequence of shows to which a user has listened.
    \item \textbf{Topic Constrained}: we constrain the data such that each of the shows in a sequence to be retained share a common topic. 
\end{itemize}

When we constrain user sequences to be within single topics we can split a user's listening history into multiple sequences each contained to a topic, and offer them a set of recommendations for each one. 

\paragraph{Embedding Styles}

We experiment with two methods of forming embeddings for the shows. In the first, as mentioned above, we use DistMult to generate embeddings of the podcast knowledge graph. These embeddings incorporate high level semantic connections between shows based on the Wikipedia pages that can be with the show descriptions (i.e. a show discussing American football may share relations with with the Wikipedia entries for famous football players) \cite{yang2014embedding}. The knowledge graph provides a much finer resolution view of podcasts than the simple topics mentioned above; the graph draws in outside information about related entities and the connections between them, ultimately placing podcasts in this same space. 

The other embedding method is to use a Continuous Bag Of Words (CBOW) embedding trained on historical listening data \citep{mikolov2013efficient}. We train the CBOW embedding model by taking sequences of
The CBOW embeddings provide embeddings that are consistent with user listening behavior, since they are trained using historical data, however they do not incorporate any higher level information about the relationship between shows that the knowledge graph provides.

\paragraph{Results and Comparisons}

\begin{figure}
    \centering
  \begin{subfigure}[b]{0.35\textwidth}
    \includegraphics[width=\textwidth]{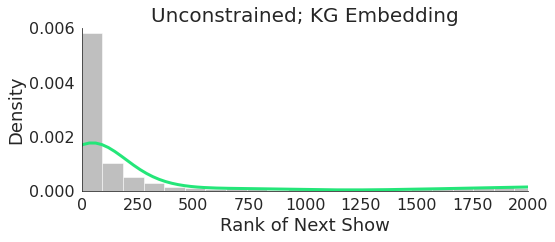}
  \end{subfigure}
  \begin{subfigure}[b]{0.35\textwidth}
    \includegraphics[width=\textwidth]{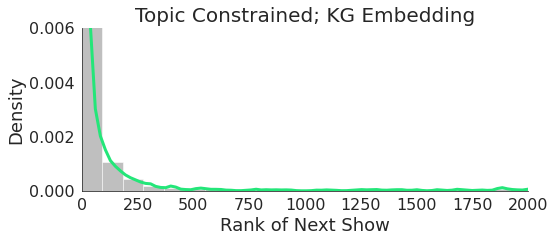}
  \end{subfigure}\\
  \begin{subfigure}[b]{0.35\textwidth}
    \includegraphics[width=\textwidth]{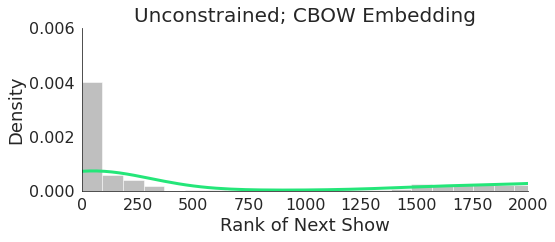}
  \end{subfigure}
  \begin{subfigure}[b]{0.35\textwidth}
    \includegraphics[width=\textwidth]{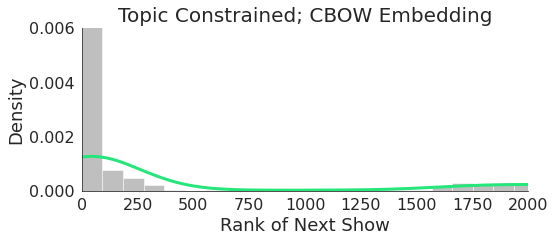}
  \end{subfigure}
  \caption{\textbf{Top}: Knowledge graph embeddings, \textbf{Bottom}: CBOW embeddings, \textbf{Left}: Unconstrained sequences of data, \textbf{Right}: Topic constrained sequences of data. We see that for all forms of data the knowledge graph outperforms a CBOW embedding, and that among all options constraining sequences of data to be within a single topic leads to the best performance.}
  \label{fig: cbow-kg}
\end{figure}

Figure \ref{fig: cbow-kg} and Table \ref{table: cbow-kg} provide comparisons for each of the embedding options and each of the data subsetting options given above. For these results we use sequences of $2$ shows and compare recommendations made to the true $3^{rd}$ show listened to by each user. We find that the inclusion of the semantic information provided by the knowledge graph greatly increases the success of the recommendations made. In addition to the inclusion of the knowledge graph, making recommendations for users within a single topic to which they have listened leads to much more successful and consistent recommendations. 

\begin{table}
\begin{minipage}{0.45\textwidth}
\centering
\begin{tabular}{ccc}
 & CBOW & KG\\
\hline \hline
Unconstrained & $0.2595$ & $0.4040$ \\
\hline
Topic & $0.3740$ & $\mathbf{0.4136}$ \\
 \end{tabular}
\end{minipage}
\begin{minipage}{0.45\textwidth}
\centering
\begin{tabular}{ccc}
 & CBOW & KG\\
\hline \hline
Unconstrained & $0.1354$ & $\mathbf{0.2069}$ \\
\hline
Topic & $0.1539$ & $0.1538$ \\
 \end{tabular}
\end{minipage}
\caption{\textbf{Left}: Success at $20$ and \textbf{Right}: Mean reciprocal rank for predictions made using CBOW and Knowledge Graph (KG) embeddings. The first row is from models trained on unconstrained sequences of user data, and the second in sequences constrained to be within a single topic. For both metrics, the highest performing model uses KG embeddings, and since our primary concern is success (rather than MRR) we favor the topic constrained and knowledge graph informed model.}
\label{table: cbow-kg}
\end{table}

\subsection{Listening Happens Locally}\label{sec: sequence-length}

We can consider the locality of user listening in either the time domain (short windows of time) or in the podcast graph (short sequences of shows).
In this section we show that by constraining our data to shorter sequences of shows aggregated over shorter time periods we can make more accurate recommendations. This section is central to our findings that podcast consumption is driven by \emph{local} behavior and the most meaningful signals can be found in a user's most recent interactions with the platform.

\begin{figure}
    \includegraphics[scale=0.3]{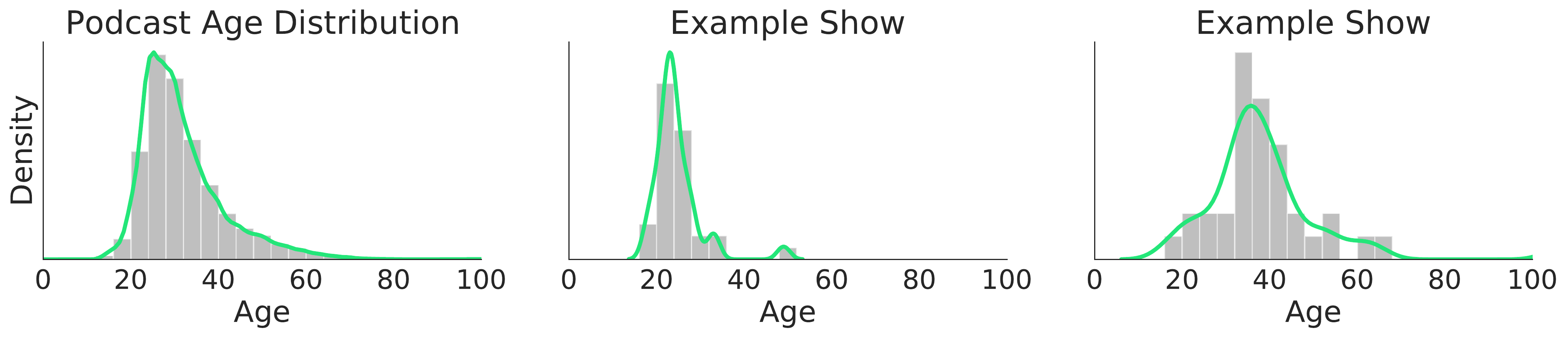}
  \caption{\textbf{Left:} the overall age distribution of listeners, \textbf{Right:} the age distributions of two example shows. We see that the age distribution of listeners to certain shows deviate significantly from the overall age distribution of listeners, indicating that a user's age can aid in informing what shows they may be more likely to listen to.}
  \label{fig: age-dist}
\end{figure}

We additionally explore one final dimension of locality: age. In Figure \ref{fig: age-dist} we see that the ages of users that listen to a given podcast may be very different from another show or from the age distribution of all listeners. 
For this reason we subset the data into bins based on the user's age, with the bins being: under $20$, $20-29$, $30-39$, $40-49$, $50-59$, and $60+$.

Below we compare performance based on the amount of time over which the data were sourced from along with the change in performance in only considering data from users aged $20-29$. For clarity of presentation we give results for this age bracket only, however in general we find that success of recommendations is overall higher when a model is specifically tailored to a specific age bracket.

\paragraph{Locality in Time and Age}

We find that constraining ourselves to only considering user behavior over short time periods gives the best performance. We compare the performance of our framework when trained on data sourced over $1$, $2$, $4$, and $6$ weeks leading up to early June, finding that the most accurate recommendations can be made when considering only user behavior over short timescales. 
The results are given in Table \ref{table: short-time}, which shows that the best performance is obtained when data are taken over the shortest time period possible. 
This effect is true for when we restrict on age or not, however the best performance is found when recommending shows for users of a specified age bracket. Moving forward all data are sourced from streams occurring over a single week by users aged $20-29$.

\begin{table}
 \begin{varwidth}{0.55\linewidth}
    \centering
    \begin{tabular}{c c c c c c}
         & Weeks of Data & $1$ & $2$ & $4$ & $6$ \\
         \hline \hline
         \multirow{2}{*}{MRR} & All Ages & $0.1538$ & $0.1409$ & $0.1320$ & $0.1130$\\
         & Age $20-29$ & $\mathbf{0.1632}$ & $0.1393$ & $0.1238$ & $0.1269$\\
         \hline
         \multirow{2}{*}{Success at $20$} & All Ages & $0.4135$ & $0.4030$ & $0.3505$ & $0.3125$\\
         & Age $20-29$ & $\mathbf{0.4515}$ & $0.4060$ & $0.3570$ & $0.3640$\\
         \hline
    \end{tabular}
     \caption{A comparison of MRR and success at $20$ for models trained using data taken from different lengths of time. The results clearly indicate a boost in performance by considering only data taken over a short period of time, with the best performance achieved by looking only over a single week and restricting to a single age bracket.}
    \label{table: short-time}
    \end{varwidth}
      \hfill
  \begin{minipage}{0.4\linewidth}
    \centering
    \includegraphics[width=40mm]{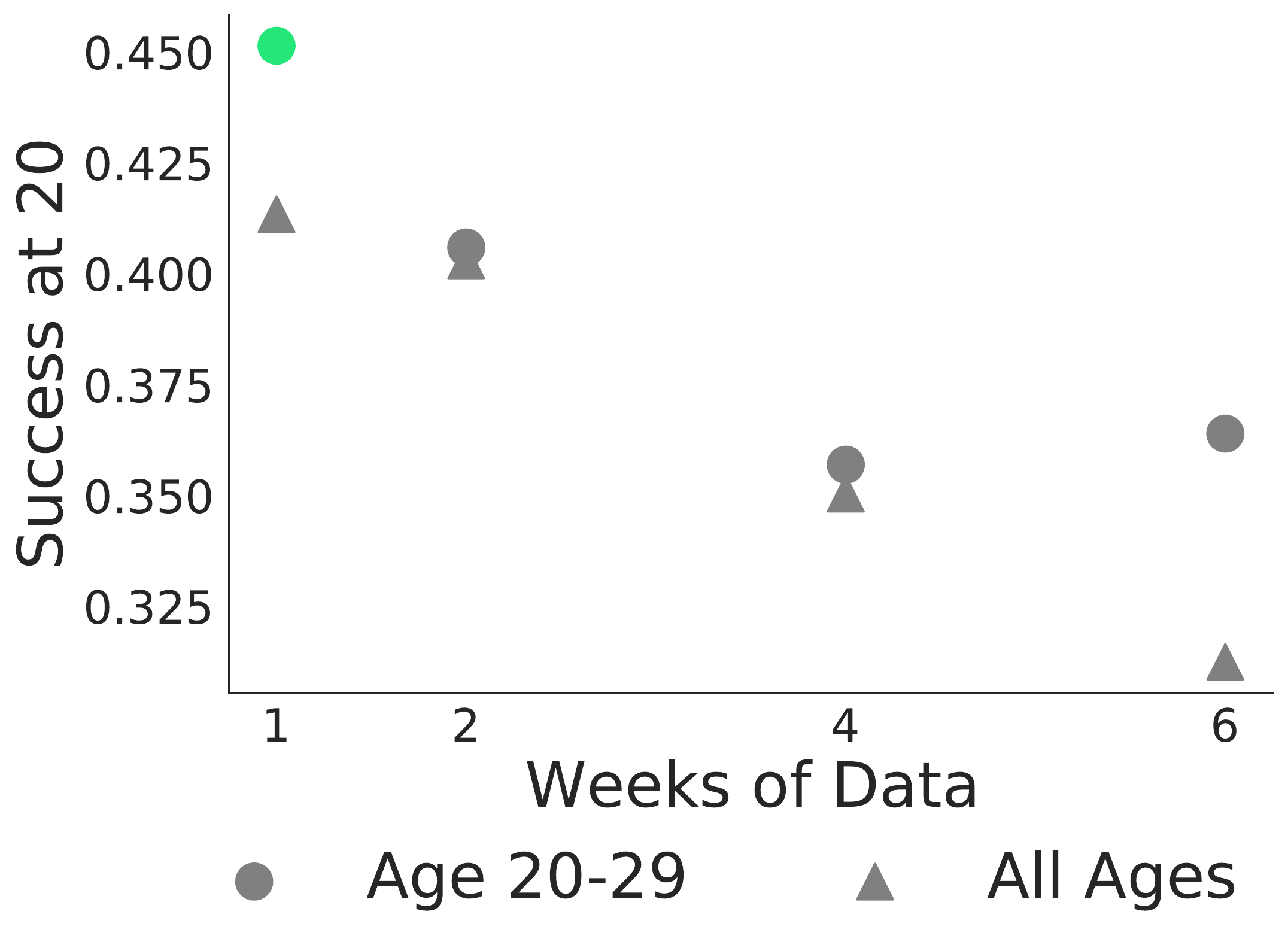}
    \captionof{figure}{Success at $20$ as a function of both the time period over which data were sourced, as well as whether or not data are subset based on age. 
    }
    \label{fig: short-time}
  \end{minipage}
\end{table}

\paragraph{Locality in the Podcast Graph}

To detect locality in the space of podcasts we consider the number of previous observations to use in order to make recommendations.
We find that using shorter sequences leads to high accuracy in recommendations, as shown in Table \ref{table: short-sequences} and Figure \ref{fig: short-time}. Again we compare the mean success of recommendation and MRR, finding that using $2$ shows to predict the $3^{rd}$ gives better accuracy than predicting the last show in longer sequences. 

\begin{table}
    \centering
    \begin{tabular}{c c c c c}
        Input Sequence Length & $2$ & $3$ & $4$ & $5$ \\
         \hline \hline
         MRR & $\mathbf{0.1605}$ & $0.1254$ & $0.1006$ & $0.0687$\\
         \hline
         Success at $20$ & $\mathbf{0.4439}$ & $0.3639$ & $0.3094$ & $0.2350$\\
         \hline
    \end{tabular}
    \caption{Comparison of MRR and success at $20$ as a function of the number of input shows used to predict the next. In both measurements the performance monotonically decreases as sequences get longer.}
    \label{table: short-sequences}
\end{table}

\subsection{Listening is Sequential}\label{sec: sequential}

As a final point of validation, we return to one of our initial assumptions and show that user sequences are indeed sequential in nature, and that the inductive biases provided by the recurrent structure in the model lead to notable accuracy gains in recommendation.
This verification of sequential structure in the data is accomplished in two ways. In the first experiment we compare our model when trained on actual data taken from user histories against the same model trained on data in which users' data has been shuffled to be out of order. In the second experiment we compare our model against a deep model with similar structure but with the recurrent structure removed, treating the inputs as a ``bag of shows" rather than as a sequence.

\paragraph{Shuffled Data}

To demonstrate that there are sequential patterns in users' behavior we test our model trained on true user data and on data in which the user sequences have been shuffled. 
Specifically we are using two previous shows consumed by a user to predict the third in the sequence (all within same topic). 
In the shuffled data we take the dataset of sequences and randomly permute the ordering of shows, generating a new set of both input sequences and target shows. 

From Section \ref{sec: structure} we see that the best performing model achieves a success at $20$ of $0.4515$. By shuffling the data we see a $17\%$ decrease in performance to a success at $20$ of $0.3774$. This results indicate that the sequences of shows consumed by users are meaningful, and that in rearranged order there is less structure to be used for making recommendations.




\paragraph{Non-Sequential Model}

Throughout this work we have made the assumption that there is a sequential nature to the way in which podcasts are consumed by users. As a final validation for our approach we remove that assumption and experiment with the accuracy that can be achieved by treating the shows a user has listened to as a non-ordered set of inputs. 

To accomplish this we use a three layer deep neural network with layers that are $512$, $1024$, and $1024$ units wide respectively. This is similar in size to the feed-forward portion of the recurrent model used in previous sections and was the highest performing architecture found in experimentation. We attempt to predict the last show in a sequence of listening by using the previous $1$, $2$, $3$, and $4$ shows, passing the inputs into the network as concatenated vectors, rather than retaining their sequential structure. Table \ref{table: nonsequential} shows the clear drop in recommendation accuracy resulting from a nonsequential treatment of the inputs. Interestingly, however, we see the same trend in success at $20$ and MRR as shown in Section \ref{sec: sequence-length} where longer sequences of inputs tend to perform worse.

\begin{table}
    \centering
    \begin{tabular}{c | c | c c c c}
         & Sequential & \multicolumn{4}{c}{Nonsequential}\\
        No. Previous Shows & $2$ & $1$ & $2$ & $3$ & $4$\\
        \hline \hline
        MRR & $\mathbf{0.1632}$ & $0.1028$ & $0.0779$ & $0.0830$ & $0.0886$ \\
        Success at $20$ & $\mathbf{0.4515}$ & $0.3544$ & $0.2456$ & $0.2504$ & $0.2638$
    \end{tabular}
    \caption{A comparison of our method (sequential) against treating the previous shows as nonsequential inputs to a feed-forward neural network. In experimenting with architectures of differing sizes and capacities, the best performance we were able to attain is still poor in comparison to the sequential approach we have introduced here.}
    \label{table: nonsequential}
\end{table}

\section{Conclusion}

We demonstrate that with the right inductive biases and the right controls on data, successful and meaningful recommendations for podcasts can be made using only short sequences of user listening histories. Imbuing our models with high level semantic information by coupling historical data with the rich structure of the knowledge graphs leads to gains over more classical and exclusively data-based approaches.
Using recurrent neural networks to view users as moving through the space of podcasts sequentially, while sourcing data taken from short periods of time provides the right structure in which accurate recommendations can be made.
Experimental evidence is provided to validate that the modeling choices made are valid, comparing against both recommendations made using non-sequential models and those made using the framework outlined here but with fewer restrictions on the data sourcing process.

\bibliographystyle{ACM-Reference-Format}
\nocite{*}
\bibliography{refs}

\end{document}